\documentclass[letterpaper]{article} % DO NOT CHANGE THIS
\usepackage{aaai2026}  % DO NOT CHANGE THIS
\usepackage{times}  % DO NOT CHANGE THIS
\usepackage{helvet}  % DO NOT CHANGE THIS
\usepackage{courier}  % DO NOT CHANGE THIS
\usepackage[hyphens]{url}  % DO NOT CHANGE THIS
\usepackage{graphicx} % DO NOT CHANGE THIS
\usepackage{natbib}  % DO NOT CHANGE THIS AND DO NOT ADD ANY OPTIONS TO IT
\usepackage{caption} % DO NOT CHANGE THIS AND DO NOT ADD ANY OPTIONS TO IT
\usepackage{algorithm}
\usepackage{algorithmic}
\usepackage{amsmath,amsfonts}
\usepackage{booktabs}
\def\eg{\emph{e.g.}~}
\def\ie{\emph{i.e.}~}

\usepackage{newfloat}
\usepackage{listings}
\DeclareCaptionStyle{ruled}{labelfont=normalfont,labelsep=colon,strut=off} % DO NOT CHANGE THIS
\floatstyle{ruled}
\newfloat{listing}{tb}{lst}{}
\floatname{listing}{Listing}
\title{Diffusion Implicit Policy for Unpaired Scene-aware Motion Synthesis}
\author{
    Jingyu Gong\textsuperscript{\rm 1, \rm 2, \rm 3}, Chong Zhang\textsuperscript{\rm 1}, Fengqi Liu\textsuperscript{\rm 4}, Ke Fan\textsuperscript{\rm 4}, Qianyu Zhou\textsuperscript{\rm 5}, Xin Tan\textsuperscript{\rm 1,2}, Zhizhong Zhang\textsuperscript{\rm 1}, Yuan Xie\textsuperscript{\rm 1,2}\thanks{Corresponding Author.}\\
}
\affiliations{
    \textsuperscript{\rm 1}School of Computer Science and Technology, East China Normal University, Shanghai, China\\
    \textsuperscript{\rm 2}Chongqing Key Laboratory of Precision Optics, Chongqing Institute of East China Normal University, Chongqing, China\\
    \textsuperscript{\rm 3}Shanghai Key Laboratory of Computer Software Evaluating and Testing, Shanghai, China\\
    \textsuperscript{\rm 4}School of Computer Science, Shanghai Jiao Tong University, Shanghai, China\\
    \textsuperscript{\rm 5}College of Computer Science and Technology, Jilin University, Jilin, China\\
    \{jygong, zzzhang, xtan, yxie\}@cs.ecnu.edu.cn, 51265901052@stu.ecnu.edu.cn,\\
    \{liufengqi, slipperyfrank\}@sjtu.edu.cn, zhouqianyu@jlu.edu.cn
}

\usepackage{bibentry}
\begin{document}

\maketitle

\begin{abstract}
Scene-aware motion synthesis has been widely researched recently due to its numerous applications. Prevailing methods rely heavily on paired motion-scene data, while it is difficult to generalize to diverse scenes when trained only on a few specific ones. Thus, we propose a unified framework, termed Diffusion Implicit Policy (DIP), for scene-aware motion synthesis, where paired motion-scene data are no longer necessary. In this paper, we disentangle human-scene interaction from motion synthesis during training, and then introduce an interaction-based implicit policy into motion diffusion during inference. Synthesized motion can be derived through iterative diffusion denoising and implicit policy optimization, thus motion naturalness and interaction plausibility can be maintained simultaneously. For long-term motion synthesis, we introduce motion blending in joint rotation power space. The proposed method is evaluated on synthesized scenes with ShapeNet furniture, and real scenes from PROX and Replica. Results show that our framework presents better motion naturalness and interaction plausibility than cutting-edge methods. This also indicates the feasibility of utilizing the DIP for motion synthesis in more general tasks and versatile scenes. Code will be publicly available at https://github.com/jingyugong/DIP.
\end{abstract}

%\begin{links}
    %\link{Code}{https://github.com/jingyugong/DIP}
    %\link{Datasets}{https://aaai.org/example/datasets}
    %\link{Extended version}{https://aaai.org/example/extended-version}
%\end{links}

\section{Introduction}
\label{sec:intro}

\begin{figure}[t!]
    \centering
    \includegraphics[width=\linewidth]{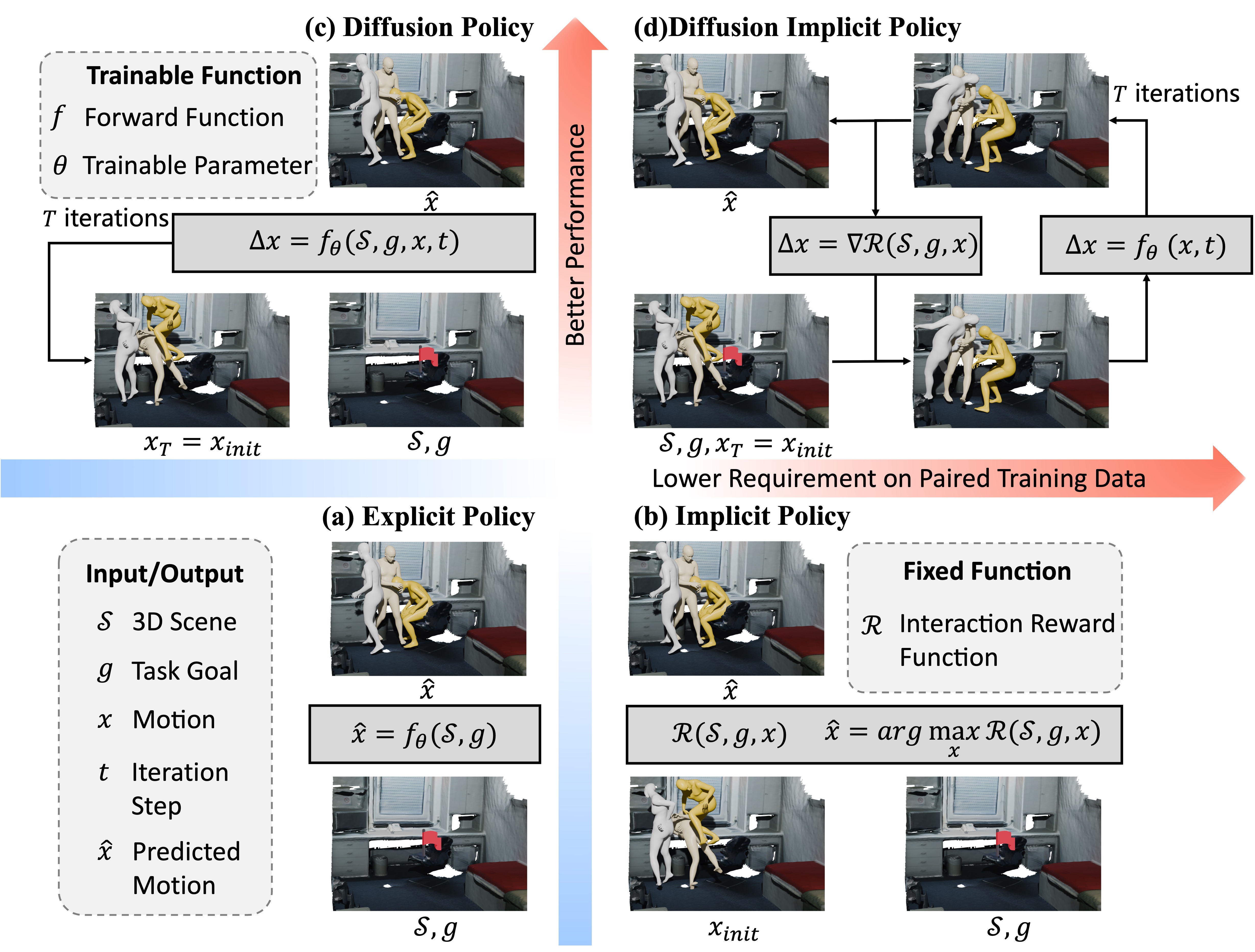}
    \caption{Policy learning frameworks. (a) Explicit policy is trained with paired motion-scene data. (b) Implicit policy optimizes the motion from random initialization. (c) Diffusion policy gradually denoise the motion. (d) Our diffusion implicit policy iteratively denoises and optimizes the motion to ensure motion naturalness, diversity, interaction plausibility without need for any paired motion-scene data.}
    \label{fig:intro}
\end{figure}

Synthesizing human motion in real 3D scenes has attracted significant attention in recent years~\cite{cao2020long,wang2021synthesizing,wang2022towards,zhao2023synthesizing,hu2024hoimotion}, due to its wide applications in scene simulation, digital human animation, and virtual/augmented reality.

%Motion synthesis is a long-existing problem in artificial intelligence and computer graphics~\cite{clavet2016motion,holden2017phase,starke2019neural,tevet2023human}. Motion matching~\cite{clavet2016motion} and MDM~\cite{tevet2023human} made good utilization of the large amounts of captured motion data to generate desired motions with satisfying performance. 

Thanks to learning-based 3D perception~\cite{qi2017pointnet,zhao2021point}, pioneers~\cite{cao2020long,wang2021synthesizing,zhao2023synthesizing} have attempted to synthesize motion in scenes with feasible human-scene interaction.
%On the other side, some prevailing works~\cite{zhang2022wanderings,zhao2023synthesizing} utilized motion data to learn a motion latent space and trained a scene-aware policy network to give plausible latent motion where pseudo motions in the scene are used for training.
However, in previous works, paired motion-scene data are required to learn scene-aware motion policies. The majority of prevailing methods~\cite{starke2019neural,zhao2023synthesizing} learn \textbf{Explicit Policies} to directly predict the desired motion based on current states and goals (Fig.~\ref{fig:intro} (a)). Some of them~\cite{wang2021synthesizing,wang2022towards} utilized a second-stage \textbf{Implicit Policy} optimization but sacrifice the motion naturalness for interaction plausibility (Fig.~\ref{fig:intro} (b)). Recent works~\cite{huang2023diffusion, wang2024move} utilized conditional \textbf{Diffusion Policies} to achieve better performance (Fig.~\ref{fig:intro} (c)), where massive paired motion-scene data is also necessary.

In fact, human motion data~\cite{mahmood2019amass,lin2023motion,ren2023lidar} is far more abundant than paired motion-scene data~\cite{hassan2019resolving,wang2022humanise}. Motion synthesis that relies heavily on paired data will inevitably suffer from limited diversity. Meanwhile, the generalization ability is hard to guarantee when trained on limited scenes and applied to various other scenes. %Thus, it will be a better choice to fully disentangle the learning of motion synthesis and human-scene interaction.

Based on this observation, we propose a unified framework, termed \textbf{Diffusion Implicit Policy (DIP)} (Fig.~\ref{fig:intro} (d)), which disentangles human-scene interaction from motion synthesis during training and then integrate motion denoising with implicit policy optimization during inference. In this way, paired motion-scene data is no longer necessary, and motion naturalness and interaction plausibility can be ensured simultaneously for scene-aware motion synthesis.

In the DIP, a motion diffusion model is employed to make the synthesized motion more and more natural throughout the entire denoising process(Fig.~\ref{fig:framework} (b)). We equipped the diffusion model with a ControlNet~\cite{zhang2023adding} branch to provide keyframe joint hints for historical motion and future goals. Following previous works~\cite{tevet2023human,xie2024omnicontrol}, the diffusion model is designed to predict the original motion at each denoising step and then sample the denoised motion from a normal distribution accordingly. Thus, we can well utilize the stochastic process to pursue plausible human-scene interactions. Specifically, interaction-based reward functions are designed to assess the consistency between motions and scenes. These reward functions are used as implicit policy to optimize the sampling distribution, ensuring that the sampled denoised motion corresponds better to the 3D scenes at each denoising step. As the denoising process can also be treated as optimization for motion naturalness, the entire scene-aware motion synthesis can be framed as an optimization problem to pursue both motion naturalness and interaction plausibility.

To synthesize reasonable motion in 3D scenes, we first train a motion diffusion model conditioned on actions and keyframe joints, which can be derived from motion itself. Furthermore, we design various reward functions to score motion naturalness and interaction plausibility. These rewards will optimize the sample distribution during motion denoising. We choose to adjust the centroid of the distribution in a GAN inversion manner, applying these reward functions to the outputs of the diffusion model at the centroid rather than directly to the centroid itself.
In this way, the proposed method can identify a better intermediate noised motion with higher motion naturalness and interaction plausibility in the final synthesized motions.

In addition, for long-term motion synthesis involving multiple tasks, we need to take historical motion as constrain when synthesizing future motion. To maintain continuity between historical and future motions, we employ a time-variant motion blending, where we interpolate the rotation matrix in the power space. Thus far, the proposed framework can synthesize long-term motion in general scenes without any training on paired motion-scene data.

For performance evaluation, we use scenes cluttered with furniture from ShapeNet~\cite{chang2015shapenet} to assess the ability on human-object interaction. We also take PROX~\cite{hassan2019resolving} and Replica~\cite{straub2019replica} to demonstrate the generalization ability in scene-aware motion synthesis. We compared the proposed method with prevailing works based on physical and perceptual scores. Comprehensive experiments support our claims and indicate that the synthesized motion produced by the proposed method demonstrates better performance.

Our main contributions can be summarized as follows: (1) We propose a brand-new framework, termed Diffusion Implicit Policy, for unpaired scene-aware motion synthesis. In this framework, we disentangle human-scene interaction from motion synthesis during training and transform scene-aware motion synthesis into a joint optimization problem, where motion naturalness and interaction plausibility are ensured by iterative diffusion denoising and implicit policy optimization. (2) We propose to adjust the centroid of the sampling distribution during denoising process in a GAN Inversion manner for higher interaction plausibility, where the motion representation is designed to be fully differentiable with respect to the human mesh and joints. (3) We design to generate new motion based on historical constrains via inpainting and blend the motion in the power space of the rotation matrix using time-variant coefficients to synthesize long-term motion for multiple subsequent tasks.

\section{Related Work}
\label{sec:related}
\noindent\textbf{Human Scene Interaction.} Generating realistic and plausible human-scene interactions has been widely explored~\cite{savva2014scenegrok,savva2016pigraphs,zhang2020place,zhang2020generating,zhao2022compositional}. 
PLACE~\cite{zhang2020place} modeled the proximity based on the distance between human body and 3D scene during interaction synthesis.
%PSI~\cite{zhang2020generating} generated human bodies in 3D scenes conditioned on scene semantics and a depth map. 
%Wang $\it{et~al.}$~\cite{wang2021synthesizing} generated a human body in scene point cloud with pre-defined translation and orientation.
POSA~\cite{hassan2021populating} designed a contact feature map for the human body, indicating the contact and semantic information for each vertex in the human mesh. 
COINS~\cite{zhao2022compositional} encoded the human body and 3D objects into a shared feature space and synthesizes diverse compositional interactions. Narrator~\cite{xuan2023narrator} modeled the correlation of 3D scene and text based on a scene graph for interaction generation.

Inspired by the optimization stage in static human-scene interaction, we design interaction-based reward functions as an implicit policy for scene-aware motion synthesis.

\noindent\textbf{Motion Synthesis.}
Motion synthesis has been studied for a significant period~\cite{clavet2016motion,holden2017phase,starke2019neural,wang2019combining,guo2020action2motion,petrovich21actor,guo2022generating,wu2024doodle}. 
%This topic has been researched conditioned on various signals, including motion prefixes~\cite{Mao_Liu_Salzmann_Li_2019}, actions~\cite{guo2020action2motion, petrovich21actor, xu2023actformer}, music~\cite{Gong_2023_ICCV,tseng2022edge}, and text~\cite{petrovich2022temos, guo2022generating, tevet2023human}. 
%Action class was widely utilized as the condition to synthesize realistic and diverse motions with desired action state. 
%Action2Motion~\cite{guo2020action2motion} employed a recurrent C-VAE for motion creation. ACTOR~\cite{petrovich21actor} encoded the entire motion sequence into a latent feature space, significantly reducing the accumulative error. 
%Furthermore, some works explored generating human motions under the guidance of natural language. 
TEMOS~\cite{petrovich2022temos} learned a shared latent space for motion and text alignment. 
MDM~\cite{tevet2023human} and MotionDiffuse~\cite{zhang2024motiondiffuse} introduced diffusion model into motion synthesis. Subsequent works~\cite{chen2023executing, xie2023towards, zhang2023remodiffuse, dai2024motionlcm, zhang2023finemogen, karunratanakul2023gmd, xie2024omnicontrol} further improved the controllability and quality of the generated results.% through database retrieval, spatial control, fine-grained captioning.

Thanks to the advancements in motion synthesis, we follow the MDM~\cite{tevet2023human} and extend it to scene-aware motion synthesis, using interaction-based reward functions as an implicit policy.

\noindent\textbf{Scene-Aware Motion Synthesis.} Synthesizing human motion in various scenes has garnered much attention recently~\cite{hassan2021stochastic,wang2022towards,zhang2022couch,hassan2023synthesizing,mullen2023placing,rempe2023trace,mir2024generating,cen2024generating,jiang2024scaling,yi2024generating,liu2024revisit}. 
Wang $\it{et~al.}$~\cite{wang2021synthesizing} utilized the PointNet~\cite{qi2017pointnet} to provide scene feature and optimized the entire motion after generation. 
%Wang $\it{et~al.}$~\cite{wang2022towards} introduced three levels of diversity for scene-aware motion synthesis. 
%SAMP~\cite{hassan2021stochastic} utilized a mixture of expert networks to first predict the action and then generate the motion. 
GAMMA~\cite{zhang2022wanderings} and DIMOS~\cite{zhao2023synthesizing} learned a latent space for natural motion, and a policy network was trained over the latent motion space. %TRACE~\cite{rempe2023trace} generated a trajectory using spatial guidance and then utilized a policy network for locomotion synthesis.  %PAAK~\cite{mullen2023placing} placed human motion in scenes according to keyframe interactions. 
SceneDiffuser~\cite{huang2023diffusion} proposed a scene-conditioned diffuser accompanied by a learning-based optimizer and planner. LAMA~\cite{lee2023locomotion} introduced a test-time optimization stage for controller network via reinforcement learning to predict the action cues for motion matching~\cite{clavet2016motion} and modification. AMDM~\cite{wang2024move} designed a two-stage framework with a scene affordance map as an intermediate representation.% for final human motion synthesis.
%However, such an approach does not take into account that only the information around person is userful for generation, but consumes a lot of resources to encode irrelevant information in the distance. 

Compared with these methods, we propose to disentangle scene-aware motion synthesis into motion prior learning via diffusion model and implicit policy learning via interaction-based reward functions, and integrate them in a unified framework, termed Diffusion Implicit Policy.

\section{Method}
\label{sec:methods}
\subsection{Preliminary}

\noindent\textbf{Motion Representation.} For human motion, we take the SMPL-X model~\cite{pavlakos2019expressive} to represent the pose at each frame. Here, we mainly consider the global orientation represented in axis-angle $\theta_{global}\in \mathbb{R}^{3}$, joint rotation in axis-angle $\theta_{j=1:21}\in \mathbb{R}^{63}$ and the translation $\tau\in \mathbb{R}^{3}$. Accordingly, for each frame $s$, the human pose can be defined as $P_s = \{\theta_{s,global},\theta_{s,j=1:21},\tau_s\} \in \mathbb{R}^{69}$, and the synthesized motion consisting of consecutive poses can be annotated as $\hat{P}=\{\hat{P}_s\}_{s=1:S}$. The body shape $\beta\in\mathbb{R}^{10}$ and hand pose $\theta_{h}\in\mathbb{R}^{24}$ are always keep the same as initial human body for simplicity. The first $K$ joints $J=J_{1:K}\in \mathbb{R}^{K\times 3}$ and body mesh with $V$ vertices $M(\tau,\theta_{global},\beta,\theta_{j},\theta_h)\in \mathbb{R}^{V\times 3}$ are taken as auxiliary representation for human pose.

\noindent\textbf{Motion Diffusion Model.} Motion diffusion is modeled as a noising process which gradually add noise to the original motion with $S$ frames $x_0=\{P_s\}_{s=1:S}$  
\begin{equation}
    q(x_t|x_{t-1}) = \mathcal{N}(\sqrt{\alpha_t}x_{t-1},(1-\alpha_t)I),
\end{equation}
where $\mathcal{N}$ is a normal distribution and $\alpha_{t=1:T}$ are hyper-parameters. The distribution of final noised motion $x_T$ will approximate to $\mathcal{N}(0,I)$. Like MDM~\cite{tevet2023human}, we train a diffusion model to predict the original motion directly 
\begin{equation}
    \label{eq:predict_phi}
    \hat{x}_0^{\phi}=\phi(x_t,t,a),
\end{equation}
where $t$ is the time step and $a$ is the action label. For motion synthesis, the denoising procedure can be formulated as:
\begin{equation}
    \label{eq:sample_phi}
    P(x_{t-1}|\hat{x}_0^{\phi},x_t) = \mathcal{N}(\mu_t(\hat{x}_0^{\phi},x_t), \tilde{\beta}_t\mathbf{I}),
\end{equation}
where $\tilde{\beta}_t = \frac{1-\bar{\alpha}_{t-1}}{1-\bar{\alpha}_t}\beta_t$, $\beta_t = 1-\alpha_t$, $\bar{\alpha}_t=\prod_{i=1}^t \alpha_i$, and $\mu_t(\hat{x}_0^{\phi}, x_t) = \frac{\sqrt{\bar{\alpha}_{t-1}}\beta_t}{1-\bar{\alpha}_t}\hat{x}_0^{\phi} + \frac{\sqrt{\bar{\alpha}_t}(1-\bar{\alpha}_{t-1})}{1-\bar{\alpha}_t}x_t$.
Thanks to the diffusion model $\phi$, we can easily adjust $P(x_{t-1})$ via optimizing $\phi(\mu_t,t-1,a)$ to pursue higher interaction plausibility in the final synthesized motion at each denoising step.

\subsection{Overview}

\begin{figure*}
    \centering
    \includegraphics[width=\linewidth]{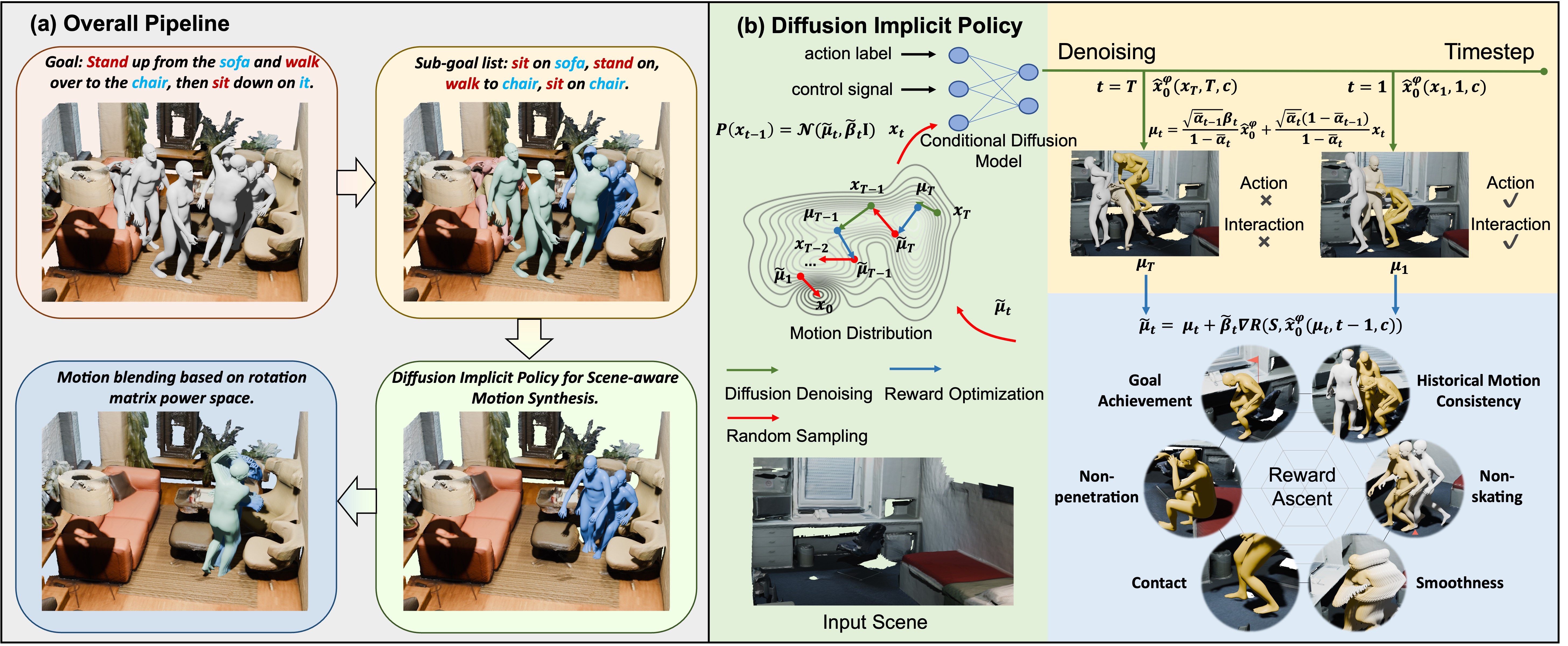}
    \vspace{-0.5cm}
    \caption{(a) indicates the overall pipeline. Any feasible command will be decomposed into sub-tasks with action-object pairs. Then, we will synthesize future motion according to current sub-task. Last, the synthesized motions will be fused into the historical motion to obtain the final long-term motion. (b) presents the Diffusion Implicit Policy (DIP). In each iteration, the denoising step will make the synthesized motion appear more \textbf{natural}), and the implicit policy optimization will endow the motion with \textbf{plausible} interaction. The random sampling step can help the framework synthesize motion with \textbf{diverse} styles. }
    \vspace{-0.5cm}
    \label{fig:framework}
\end{figure*}

In this paper, we attempt to synthesize human motion in 3D scenes given a sequence of interaction sub-tasks (Fig.~\ref{fig:framework} (a)). 

We can first decompose the command into a list of sub-task (interaction behavior and object pair) via current LLMs~\cite{achiam2023gpt,touvron2023llama}.
For each sub-task, the human may go to some place or interact with the objects in the scene (\eg, sitting on the chair). 

Given a sub-task, we will first locate the goal position using COINS~\cite{zhao2022compositional}, and then fetch reward functions for implicit policy according to current action.

We train a diffusion model conditioned on human action and keyframe joints to synthesize human motion with easy control. Later, we model the interaction-based reward functions and take them to optimize the sampling distribution of denoised motion at each denoising step. The motion prior from diffusion model and implicit policy from reward function are integrated together to synthesize scene-aware motion with desired interactions (as shown in Fig.~\ref{fig:framework} (b)).

Given historical motion, the synthesized motion should be consistent with it. Thus, we design to derive the long-term motion via a motion blending where translation are interpolated linearly and rotation are blended in the matrix power space. By now, we can synthesize long-term motion in 3D scenes when all sub-tasks are completed.

\subsection{Conditional Diffusion Model}
\label{subsec:controlnet}
\begin{figure}
    \centering
    \includegraphics[width=\linewidth]{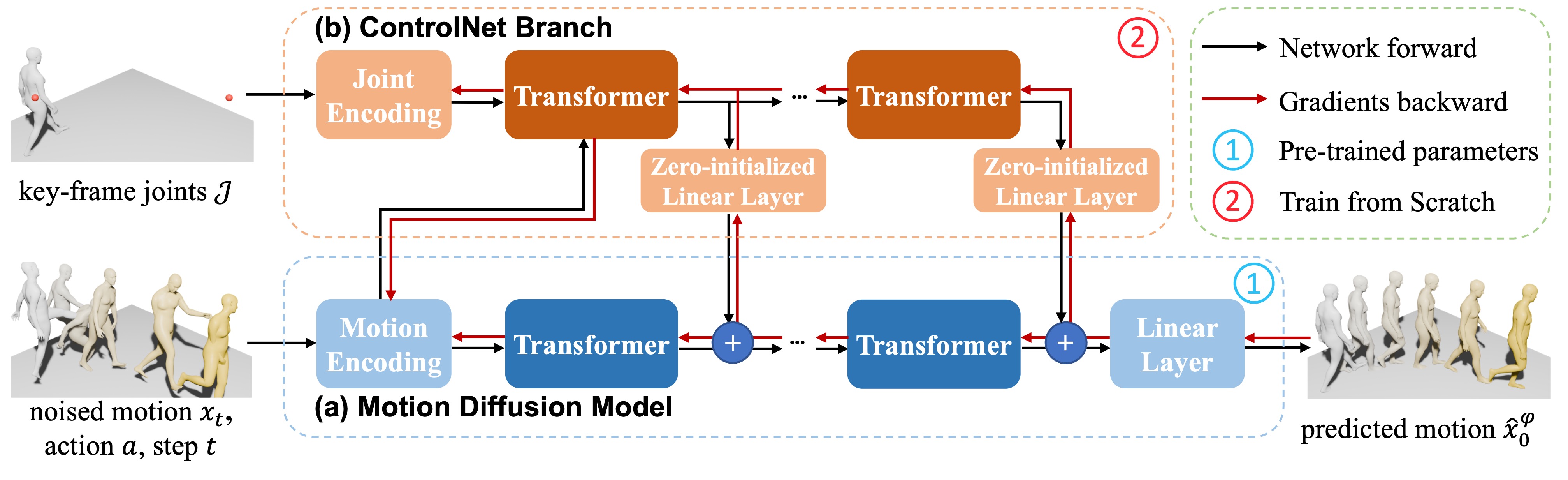}
    \vspace{-0.4cm}
    \caption{Illustration of conditional diffusion model. A diffusion model is first trained conditioned on action, and then a ControlNet branch is taken to provide keyframe joints' hint.}
    \vspace{-0.4cm}
    \label{fig:controlnet}
\end{figure}

We take $x_0\in\mathbb{R}^{S\times 69}$ consisting of human translation, orientation, and joint rotations in $S$ frames as motion representation. To simplify the formulation of reward functions in our method, we train a diffusion model $\phi$ (Fig.~\ref{fig:controlnet} (a)) to predict the original motion $\hat{x}_0^{\phi}$ for each noised $x_t$ as shown in Eq.~\ref{eq:predict_phi}.  

For any original motion $x_0$, we take the local coordinate of motion to reduce representation redundancy. Specifically, we translate and horizontally rotate the motion based on the human pose in the first frame. In the transformed motion, the orientation of the human body in the first frame lie in the $yz$-plane ($y\ge 0$), and the pelvis is positioned at the origin. After motion synthesis, the motion in scenes can be derived via simple translation and horizontal rotation.

To better maintain consistency with historical motion and achieve future goals, we also take a ControlNet branch~\cite{zhang2023adding} to provide the hint of keyframe human joints (available from motion data for training) that need to be controlled (Fig.~\ref{fig:controlnet} (b)). We choose to take the joint positions rather than joint rotation as the external hint for easier understanding of the space information. %Here, we mainly use the principle $22$ joints from the SMPL-X human body skeleton as the skeleton joint hint. 
Concretely, we will first calculate the joints' positions according to the original human motion, and then randomly select one from these joints in a few frames as the hint and others are padded with zeros, thus the input of ControlNet branch take the form of $\mathcal{J}=\{J_{s,k}\}_{s=1:S,k=1:K}$ where non-zero values provide the controlled joint position hints.

For training the ControlNet branch, all its parameters are randomly initialized, and the link layers are initialized to zero to maintain the motion synthesis capability of the main branch. Meanwhile, all parameters in original motion diffusion model $\phi$ are loaded from a pre-trained model. The controlled diffusion model $\varphi$ is also supervised to predict the origin motion $x_0=\{P_s\}_{s=1:S}$.
After fine-tuning, the controlled motion diffusion model can be formulated as 
\begin{equation}
    \hat{x}_0^\varphi = \varphi(x_t, t, a, \mathcal{J}).
\end{equation}
For simplicity, the action $a$ and joint hint $\mathcal{J}$ are termed as the condition $c$. Later, the controlled diffusion model $\varphi$ are taken as motion prior to ensure motion naturalness.

To better control the poses at specific frames, especially for the historical motion constrains, we maintain the fixed poses in an inpainting manner when $t>T_{inpaint}$. The predicted original motion $\hat{x}_0^{\varphi}$ will be updated via inpainting that can be formulated as $\hat{x}_0^\varphi=m\cdot x_m + (1-m)\cdot \hat{x}_0^{\varphi}$, where $x_m$ indicates the poses that should be maintained and $m\in \mathbb{R}^{S\times 69}$ represents the inpainting mask.

Thus far, the diffusion model can provide motion prior with explicit joint control for later DIP.

\subsection{Reward Functions for Implicit Policy}
\label{subsec:reward}
Based on the aforementioned diffusion model, we introduce a set of differentiable reward functions that score the motions within scenes and serve as an implicit policy to enhance overall performance (as shown in Fig.~\ref{fig:framework} (b) blue part). Here, we primarily focus on three key aspects in the design of reward functions: motion continuity, goal achievement, and interaction plausibility.

To promote motion continuity, we introduce a Historical Motion Consistency reward $\mathcal{R}_{his}$ to ensure continuous transition between historical and synthesized motions, alongside a Smoothness reward to prevent abrupt changes with excessive acceleration $\mathcal{R}_{acc}$. Additionally, we implement a Goal Achievement reward $\mathcal{R}_{goal}$ that encourage the human to move closer to the target. We also incorporate a Contact reward $\mathcal{R}_{cont}$ and a Non-Penetration reward $\mathcal{R}_{pene}$ to facilitate plausible human-scene interactions, as well as a Non-skating reward $\mathcal{R}_{skt}$ to anchor the vertices of contact. The total reward function for implicit policy take the form of
\begin{equation}
\label{eq:total}
\begin{array}{rcl}
    \mathcal{R}_{ip} &\hspace{-0.3cm}=&  \hspace{-0.3cm}\lambda_{his}\mathcal{R}_{his} + \lambda_{acc} \mathcal{R}_{acc} + \lambda_{goal} \mathcal{R}_{goal}\\
    & & \hspace{-0.3cm} +\lambda_{cont} \mathcal{R}_{cont} + \lambda_{pene} \mathcal{R}_{pene} + \lambda_{skt}\mathcal{R}_{skt}
\end{array}
\end{equation}
where $\lambda_{(\cdot)}$ are a series of hyper-parameters.

\subsection{Diffusion Implicit Policy}
\label{subsec:dip}
By now, we utilize a diffusion model with ControlNet branch $\varphi$ to predict $\hat{x}_0^{\varphi}$. Thus, similar to Eq.~\ref{eq:sample_phi}, we can sample the $x_{t-1}$ according to $x_t$ and $\hat{x}_0^{\varphi}$
\begin{equation}
    \label{eq:sample}
    P(x_{t-1}|\hat{x}_0^{\varphi},x_t) = \mathcal{N}(\mu_t(\hat{x}_0^{\varphi},x_t), \tilde{\beta}_t\mathbf{I}).
\end{equation}
Such denoising process (Fig.~\ref{fig:framework} (b) yellow part) can also be considered as an optimization problem based on a motion naturalness reward function $\mathcal{R}_{nat}$ which can be defined implicitly by its gradient $\nabla \mathcal{R}_{nat}(x_t) = \mu_{t}-x_t$. Meanwhile, the denoising process is also accompanied by a stochastic disturbance item (Fig.~\ref{fig:framework} (b) green part) following $\mathcal{N}(0,\tilde{\beta}_t\mathbf{I})$. 

Thus, the stochastic item can be well utilized to search for motion with higher interaction plausibility. We design the interaction-based implicit policy (Fig.~\ref{fig:framework} (b) blue part) to partly play the role of such disturbance, and scene-aware motion synthesis can be treated as a joint optimization problem which both maximize the motion naturalness and interaction plausibility in Diffusion Implicit Policy
\begin{equation}
    \hat{x}_0 = arg \max_x \mathcal{R}_{dip}(x),
\end{equation}
where
\begin{equation}
    \mathcal{R}_{dip}(x) = \mathcal{R}_{nat}(x) + \mathcal{R}_{ip}(\hat{x}_0^\varphi(x)).
\end{equation}

In order to synthesize human motion that can maximize the total reward, we integrate the implicit policy optimization into each denoising step where motion naturalness and interaction plausibility can be enhanced iteratively. 

In Eq.~\ref{eq:sample}, $x_{t-1}$ is sampled from $\mathcal{N}(\mu_t,\tilde{\beta}_t\mathbf{I})$, thus we can adjust $\mu_{t}$ (\ie the mean value of $x_{t-1}$) based on implicit policy and more suitable $x_{t-1}$ can be sampled accordingly. 
It is noteworthy, we need the final synthesized human motion $x_0$ to be consistent with the scene and achieve high reward. Thus, we propose to optimize $\mu_{t}$ through $\hat{x}_0^{\varphi}(\mu_t, t-1, c)$ rather than $\mu_t$ itself. Here, we take $t-1$ as denoising step because $\mu_t$ is the mean value of the denoised $x_{t-1}$.

Thanks to the motion representation that we take, the reward functions are fully differentiable. Meanwhile, we find that optimizing $\hat{x}_0^{\varphi}(\mu_t, t-1, c)$ shows better performance than directly modifying $\mu_{t}$ itself as previous work~\cite{xie2024omnicontrol}. That is because direct optimization over $\mu_t$ does not ensure motion continuity. On the other side, optimizing $\mu_t$ via $\hat{x}_0^{\varphi}(\mu_t, t-1, c)$ can help search a better distribution with higher interaction reward for the final synthesized motion $x_0$, and $\mu_t$ can be be adjusted as a whole in a GAN Inversion manner (given a Generator $x=G(z)$, adjust the latent code $z$ via loss $\mathcal{L}(x)$ for more desired output)~\cite{bau2019semantic,bau2019seeing}. Similarly, given the diffusion model $\hat{x}_0^{\varphi}(\mu_t, t-1, c)$, we optimize $\mu_t$ via $\mathcal{R}(\hat{x}_0^{\varphi})$ according the following formulation
\begin{equation}
    \tilde{\mu}_t = \mu_t + \tilde{\beta}_t\cdot\nabla\mathcal{R}_{ip}(\mathcal{S}, \hat{x}_0^{\varphi}(\mu_t, t-1, c)),
\end{equation}
where $\mathcal{S}$ indicate the 3D scene information, including scene semantics, SDF, and floor height. Further, $\tilde{\mu}_t$ is taken as the mean of distribution to sample $x_{t-1}$.

\subsection{Multi-Task Motion Synthesis}
\label{subsec:blending}
As for the command for multi-task motion synthesis, such as ``The person first sits on the bed, then goes to the corner of the room, and finally sits on the chair.'', we need to infer future motion ``Sit on the chair'' while maintaining continuity with previously synthesized motions ``The person first sits on the bed and then goes to the corner of the room''.

For any previous synthesized motion $\tilde{P}_{1:\tilde{S}}$, we will select the latest $H=min(\tilde{S},H_{max})$ frames as external historical constrains for future motion synthesis. Explicitly, we extract those pelvis joints $\{\tilde{J}_{s,pelvis}\}_{s=\tilde{S}-H+1:\tilde{S}}$ as trajectory hints for conditional motion diffusion. In addition, we take the body skeleton joints from the historical frames to form $\mathcal{J}=\{\tilde{J}_s\}_{s=\tilde{S}-H+1:\tilde{S}}$, and use them for pose constraints in the implicit policy optimization.

After a new round of motion synthesis, we obtain the generated motion $\{\hat{P}_s\}_{s=1:S}$. For more natural motion transition in the overlapping $H$ frames, we take a time-variant motion blending. Different from direct linear interpolation in the pose representation like priorMDM~\cite{shafir2024human}. We only utilize linear interpolation for the human translation $\check{\tau}_{1:H}$. As for human orientation and joints' rotations, we take the axis-angle representation $\bar{\theta}_{rot,s}=\tilde{\theta}_{\tilde{S}-H+s}$ and $\hat{\theta}_{rot,s}$ and convert them to the rotation matrix $\bar{M}_{rot,s}$ and $\hat{M}_{rot,s}$. The blending occur in the power space of rotation matrix which can be formulated as:
\begin{equation}
    \check{M}_{rot,s} = (\hat{M}_{rot,s}\bar{M}_{rot,s}^{-1})^{\gamma}\bar{M}_{rot,s}
\end{equation}
where $\gamma=(1:H)/(H+1)$, and $\check{\theta}_{1:H}$ is converted from $\check{M}_{rot,1:H}$. Thus, the blended pose take the form of $\check{P}_{1:H}=\{\check{\theta}_{s,global},\check{\theta}_{s,j},\check{\tau}_s\}_{s=1:H}$. The newly updated long-term motion is derived as
\begin{equation}
    \tilde{P}_{1:\tilde{S}+S-H} = \{\tilde{P}_{1:\tilde{S}-H}, \check{P}_{1:H}, \hat{P}_{H+1:S}\}.
\end{equation}
The entire motion can be synthesized in an iterative manner until all sub-tasks are completed (Fig.~\ref{fig:framework} (a)).

\section{Experiments}
\label{sec:exp}

\begin{table}
    \centering
    \caption{Evaluation of motion synthesis on locomotion task. The up/down arrows ($\uparrow$/$\downarrow$) indicate higher/lower is better. Metrics with best performance are annotated in boldface.}
    \begin{footnotesize}
    \begin{tabular}{lcccc}
        \toprule
        & time $\downarrow$ & avg. dist $\downarrow$ & contact $\uparrow$ & loco pene $\uparrow$ \\
        \midrule
        SAMP& 5.97 & 0.14 & 0.84 & 0.94 \\
        GAMMA& 3.87 & \textbf{0.03} & 0.94 & 0.94\\
        DIMOS & 6.43 & 0.04 & \textbf{0.99} & \textbf{0.95}\\
        Ours & \textbf{3.35} & \textbf{0.03} & 0.91 & \textbf{0.95} \\
        \bottomrule
    \end{tabular}
    \end{footnotesize}
    \label{tab:locomotion}
\end{table}

\begin{figure}
    \centering
    \vspace{-0.4cm}
    \includegraphics[width=\linewidth]{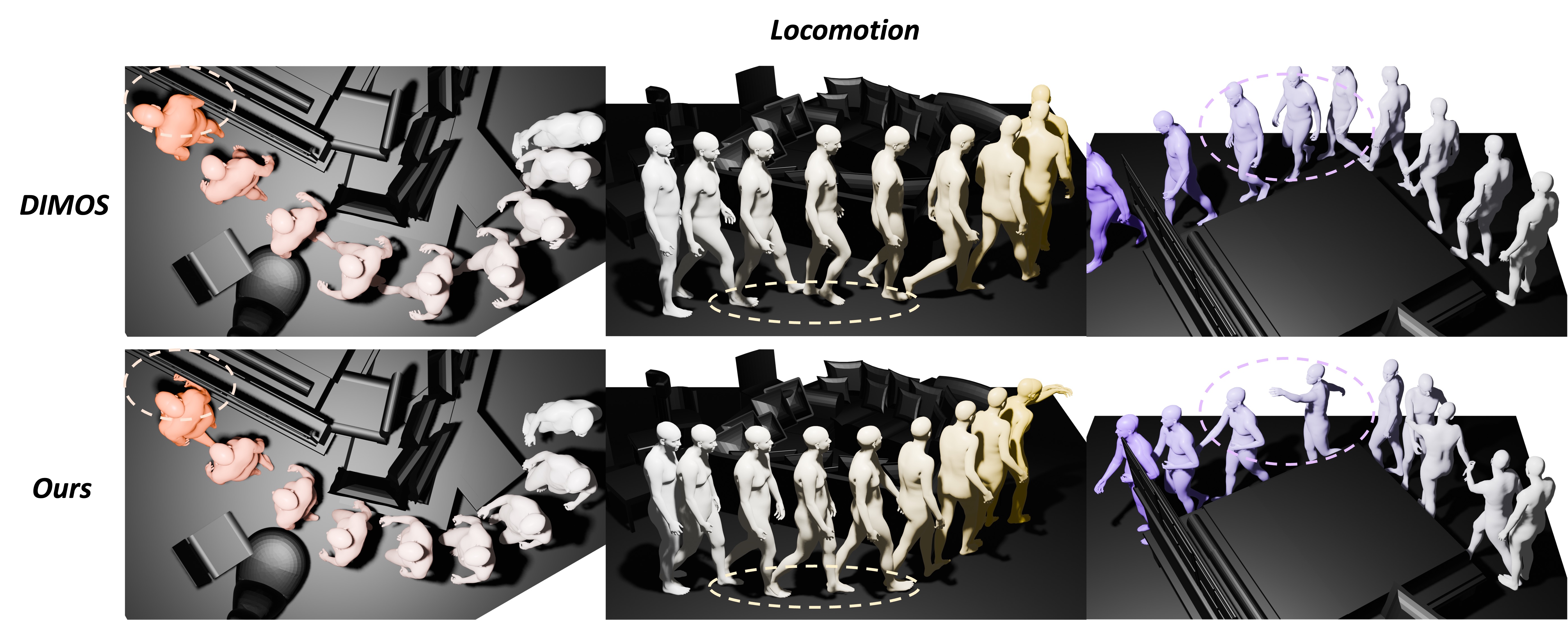}
    \vspace{-0.4cm}
    \caption{Visual results given by DIMOS and ours for locomotion task. The dashed circles indicate lower penetration, less skating and higher diversity in the synthesized motion.}
    \vspace{-0.5cm}
    \label{fig:locomotion}
\end{figure}

\begin{figure*}
    \centering
    \includegraphics[width=\linewidth]{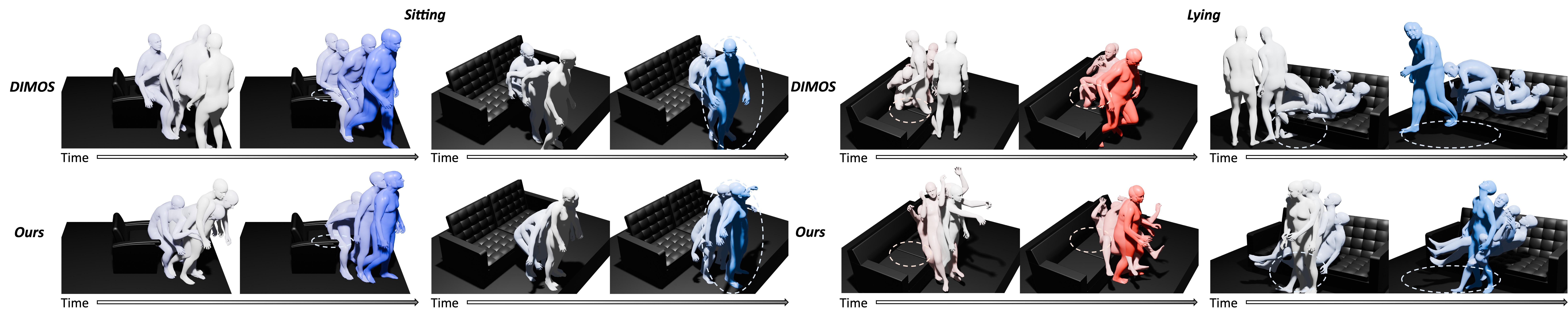}
    \vspace{-0.4cm}
    \caption{Visual results of synthesized motions given by DIMOS and our method for sitting and lying. The dashed circles indicates obvious advantages over DIMOS in less collision (col. 1,3), higher diversity (col. 2) and better foot contact (col. 4).}
    \label{fig:sit_lie}
     \vspace{-0.4cm}
\end{figure*}

\subsection{Datasets}
\label{subsec:datasets}

\noindent\textbf{Motion Datasets.}
Here, we use captured motion data with action/description labels from AMASS~\cite{mahmood2019amass} to train our model. Babel~\cite{punnakkal2021babel} provided action labels and the start/end frames for several subsets of AMASS. HumanML3D~\cite{guo2022generating} provided additional sentence annotations and start/end frames for more motion data in AMASS.

\noindent\textbf{Scene Datasets.} We evaluate the performance in both synthesized scenes and real scanned scenes. Following DIMOS~\cite{zhao2023synthesizing}, we use randomly generated scenes consisting of furniture from
ShapeNet~\cite{chang2015shapenet} to validate the performance on atomic locomotion and human-scene interaction. As for real scanned scenes from PROX~\cite{hassan2019resolving} and Replica~\cite{straub2019replica}, we take them to evaluate the performance of the pipeline in synthesizing long-term motions within scenes that involve multiple tasks. All experiments are conducted using the same model and pipeline, thus indicating the generalization ability of our method. 

\subsection{Scene Navigation}

We take the generated scenes from DIMOS~\cite{zhao2023synthesizing} for testing, where scenes are cluttered with furniture from ShapeNet~\cite{chang2015shapenet}. In this experimental setting, the human need to walk from the starting point to the target point and avoid collisions with the furniture in scenes. 

\noindent\textbf{Metrics.} We also evaluate the performance of synthesized motion for locomotion from four aspects as DIMOS~\cite{zhao2023synthesizing}, where (1) finish time measured in seconds, (2) average horizontal distance from final human body to target point measured in meters, (3) foot joint contact score 
\begin{equation}
    \label{eq:contact_score}
    s_{cont}=e^{-(|min j_z|-0.05)_{+}}\cdot e^{-(min ||j_{vel}||_2-0.075)_{+}},
\end{equation}
and (4) penetration score indicating the percentage of body vertices within the walkable area are taken as our metrics.

\noindent\textbf{Results.} We compare our method with SAMP~\cite{hassan2021stochastic}, GAMMA~\cite{zhang2022wanderings}, and DIMOS~\cite{zhao2023synthesizing} for locomotion in 3D scenes and report the results in Tab.~\ref{tab:locomotion}. The results indicate our method can achieve the shortest finish time (3.35s), closest distance (0.03m), and lowest penetration (0.95). It's noteworthy, the locomotion speed of the proposed method is much similar to that of real human than other methods. As can be seen, the performance on the contact score is inferior. We believe that is because our method focuses more on foot vertex contact, whereas the contact score calculation is based on foot joints.%the body mesh markers works better than motion representation based on axis-angle for foot stabilization.

We visualize a few examples of DIMOS and the proposed for locomotion task in Fig.~\ref{fig:locomotion}. As shown, the synthesized motion for locomotion demonstrates lower scene penetration, less skating and higher motion diversity. That is attributes to the implicit policy and the stochastic sampling procedure during denoising.

\subsection{Scene Object Interaction}
We take scenes with furniture from ShapeNet~\cite{chang2015shapenet} to evaluate the performance of the proposed method on scene object interaction, and $10$ objects ($3$ armchairs, $3$ straight chairs, $3$ sofas and $1$ L-sofa) are chosen for atomic interaction as previous work~\cite{zhao2023synthesizing}. For each scene, the human is initialized to stand in front of the interaction object and guided to interact with it and finally return to original position.

\noindent\textbf{Metrics.} We take 4 metrics to evaluate the performance, including (1) the time to finish the task measured in seconds, (2) the foot contact score mentioned in Eq.~\ref{eq:contact_score}, (3) mean human mesh vertex penetration $\sum_{v\in M}|(f_{SDF}(v))_{-}|$ over time, and (4) maximum penetration over time.

\noindent\textbf{Results.} We compare the proposed method with prevailing methods~\cite{hassan2021stochastic,zhao2023synthesizing} on human-scene interaction. The interaction tasks for sitting and lying are evaluate separately, and the results are reported in Tab.~\ref{tab:interaction}. 

\begin{table}[!ht]
    \centering
    \caption{Evaluation of motion synthesis on interaction tasks. The up/down arrows ($\uparrow$/$\downarrow$) indicate higher/lower is better. The best results are shown in boldface.}
    \begin{footnotesize}
    \begin{tabular}{lccc}
        \toprule
        & time $\downarrow$ &  pene. mean $\downarrow$ & pene. max $\downarrow$ \\
        \midrule
        SAMP sit& 8.63 & 11.91 & 45.22 \\
        DIMOS sit& 4.09 & 1.91 & 10.61\\
        Ours sit& \textbf{3.71} & \textbf{1.86} & \textbf{7.13}\\
        \midrule
        SAMP lie & 12.55 & 44.77 & 238.81 \\
        DIMOS lie & 4.20 & 9.90 & 44.61 \\
        Ours lie & \textbf{3.55} & \textbf{9.80} & \textbf{30.8} \\
        \bottomrule
    \end{tabular}
    \end{footnotesize}
    \label{tab:interaction}
\end{table}

Fig.~\ref{fig:sit_lie} visualizes the synthesized human-scene interaction given by DIMOS and our method. The visual results indicate our method presents more plausible interaction with higher human-scene contact and lower human-scene collision.

\begin{figure*}
    \centering
    \includegraphics[width=\linewidth]{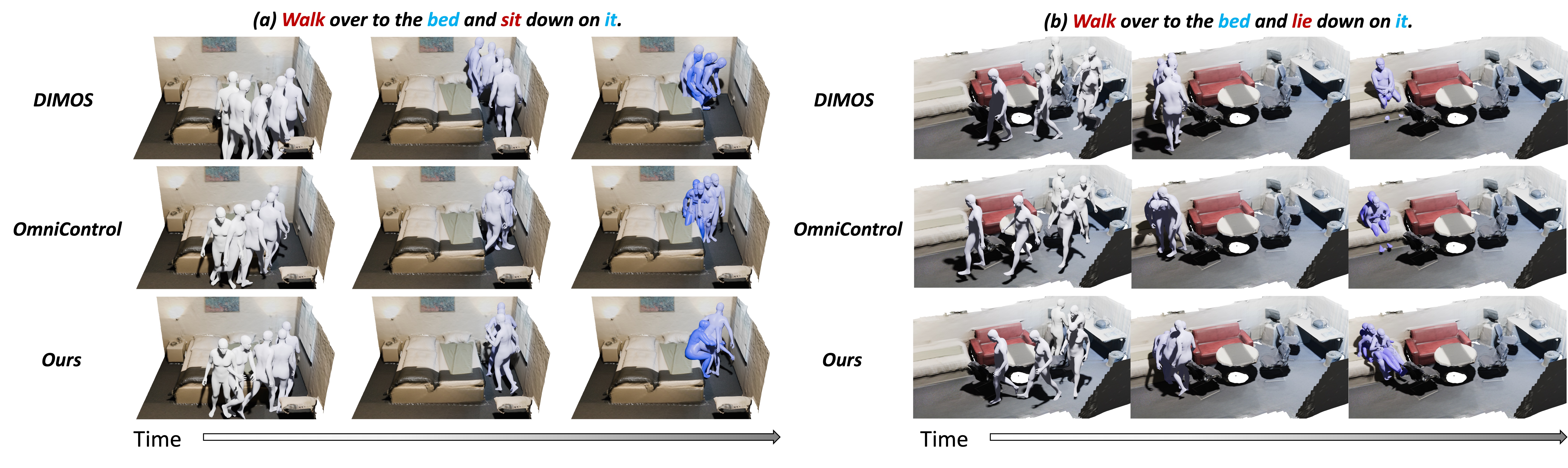}
    \vspace{-0.6cm}
    \caption{Visual comparison of different methods on motion synthesis in 3D scenes from PROX. }
    \vspace{-0.4cm}
    \label{fig:prox}
\end{figure*}

\begin{figure*}
    \centering
    \includegraphics[width=\linewidth]{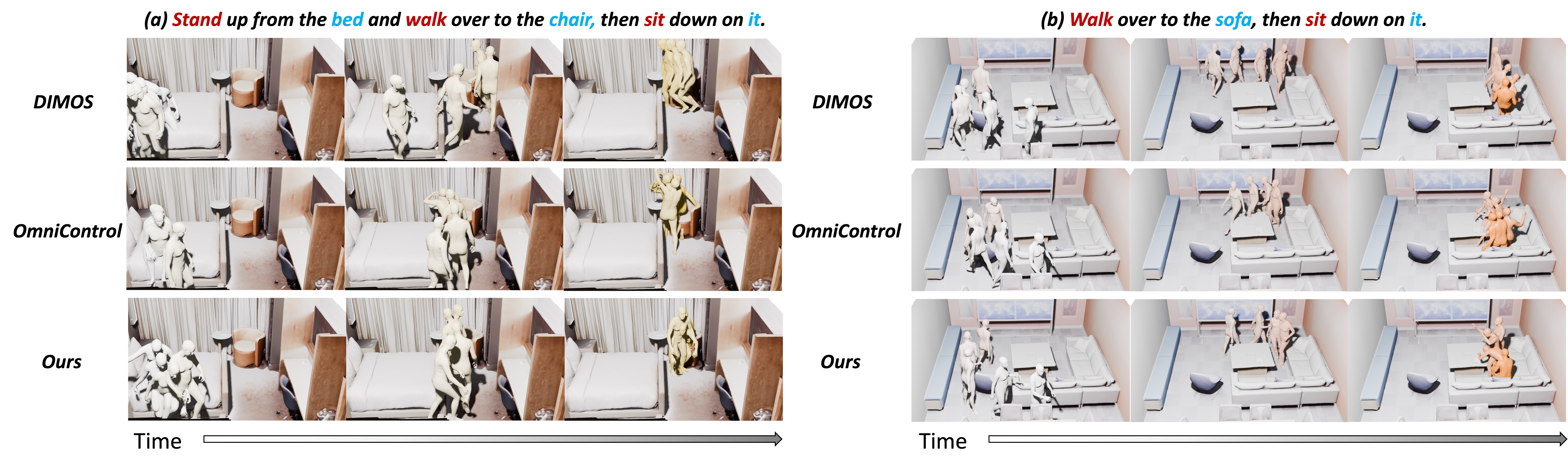}
    \vspace{-0.5cm}
    \caption{Visual results of synthesized motions given by compared methods in Replica scenes.}
    \vspace{-0.5cm}
    \label{fig:replica}
\end{figure*}

\subsection{Long-term Motion Synthesis}

For long-term motion synthesis in 3D scenes where multiple tasks are completed consecutively, objects with feasible interaction in scenes are randomly selected. We utilize COINS~\cite{zhao2022compositional} to sample the static interactions with these objects as the goals. All compared methods use the same initial state and task goals to synthesize long-term motions, ensuring a fair comparison.

\noindent\textbf{Metrics.} To comprehensively evaluate the synthesized motions, we conduct a user study where results given by different methods are directly rated by participants. We totally generate $20$ motion for each method, $10$ for scenes from PROX, and $10$ for scenes from Replica. We present motions synthesized by different methods in the same scene to participants simultaneously, and ask the them to rate the results on a scale of 1 to 5 (in increments of 0.5) from four perspectives: motion naturalness, diversity, interaction plausibility and overall performance.

\noindent\textbf{Results.} Finally, $1,200$ ratings from $15$ participants are collected for each method ($60$ results per sample from $4$ aspects) and we report the scores in Tab.~\ref{tab:user}. It can be seen, the proposed method achieve the best performance in motion diversity, interaction plausibility and overall performance even without paired motion-scene data for training. For motion naturalness, our proposed method is on par with DIMOS, as no specific designs are taken to improve motion naturalness. %that is because motion data used for training share similar distribution. 

\begin{table}[h!]
\centering
%\begin{center}
\caption{Comparison between competitive methods based on user study. Users are asked to give scores (ranging from $1$ to $5$, $\uparrow$) according to motion naturalness, diversity, interaction plausibility, and overall performance. Results on PROX/Replica are reported on the left/right respectively.}

\begin{scriptsize}
\begin{tabular}{lcccc}
\toprule
& Naturalness $\uparrow$ & Diversity $\uparrow$ & Plausibility $\uparrow$ & Overall $\uparrow$\\
\midrule
DIMOS & 2.72/\textbf{3.09} & 3.00/3.21  & 2.55/2.85 & 2.86/3.11 \\
OmniControl & 2.66/2.67 & 2.83/3.07 & 2.26/2.5 & 2.61/2.69 \\
Ours & \textbf{2.81}/3.03 & \textbf{3.34}/\textbf{3.36} & \textbf{3.15}/\textbf{3.17} & \textbf{3.17}/\textbf{3.26} \\
\bottomrule
\end{tabular}
\end{scriptsize}
%\end{center}
\label{tab:user}
\end{table}

We present the results of motion synthesis in scenes from PROX and Replica dataset respectively in Fig.~\ref{fig:prox} and Fig.~\ref{fig:replica}. It can be seen that the proposed Diffusion Implicit Policy performs better in terms of motion naturalness, interaction plausibility and motion diversity. This is due to the integration of motion denoising, implicit policy optimization, and random sampling within a unified framework.

\section{Conclusion}

In this paper, we propose a unified framework, termed Diffusion Implicit Policy, for motion synthesis in 3D scenes, where paired motion-scene data is no longer needed for training. In this framework, interaction is disentangled from motion learning during training. Motion prior from diffusion model and implicit policy from reward functions are integrated together to iteratively optimize the motion from random noise, pursuing motion naturalness, diversity and interaction plausibility simultaneously. We utilize joint hints and inpainting to ensure that keyframe poses remain consistent with the historical motion. We adjust the motion distribution centroid in a GAN Inversion manner to achieve better interaction plausibility while maintaining motion continuity. We introduce motion blending in the power space of the joint rotation matrix to ensure smooth transitions between multiple tasks for long-term motion synthesis. Comprehensive experiments on generated scenes with ShapeNet furniture, and scenes from PROX and Replica indicate the effectiveness and generalization capability. Such a promising solution can also encourages future works to learn scene-aware motion synthesis from unpaired motion and scene data.

\section*{Acknowledgements}
This work is supported by the National Natural Science Foundation of China (Grant No. 62222602, 62302167, U23A20343, 62502159), Natural Science Foundation of Shanghai (Grant No. 25ZR1402135), Shanghai Sailing Program (Grant No. 23YF1410500), Young Elite Scientists Sponsorship Program by CAST (Grant No. YESS20240780), the Chenguang Program of Shanghai Education Development Foundation and Shanghai Municipal Education Commission (Grant No. 23CGA34), Natural Science Foundation of Chongqing (Grant No. CSTB2023NSCQ-JQX0007,CSTB2023NSCQ-MSX0137, CSTB2025NSCQ-GPX0445), Open Project Program of the State Key Laboratory of CAD\&CG (Grant No. A2501), Zhejiang University.

\bibliography{aaai2026}

\appendix

\section{Implementation Details}
\label{sec:implementation}
In this section, we will provide implementation details on motion data preparation, motion coordinate transformation, reward design and motion synthesis procedure.
\subsection{Data Preparation}
We utilize the motion data from AMASS~\cite{mahmood2019amass} for training. Specifically, Babel~\cite{punnakkal2021babel} provided action labels and HumanML3D~\cite{guo2022generating} provided additional sentence annotations for several subsets of the AMASS. For these motions, we match the keywords and categorize them into three states. The keywords for action state categorization are shown in Tab.~\ref{tab:keyword}. All motions are downsampled to $40$ FPS and split into $160$-frame motion clips with a $20$-frame stride. Motion clips are all transformed according to the human pose in the first frame, where the transformed initial pose is centered (with the pelvis located at the origin) and orientation lie in $yz$-plane ($y\ge 0$). All the motion clips, along with the action labels, are used to train the motion diffusion model. Additionally, skeletons are extracted to provide joint position hints when training the ControlNet branch.

\begin{table}[h]
    \centering
    \caption{Keywords used to match the action labels or sentence annotation in Babel and HumanML3D, and the matched motions are categorized into specific actions.}
    \begin{tabular}{lccc}
        \toprule
        Action & locomotion & sit & lie\\
        \midrule
        Keywords& walk, turn, jog, run & sit & lie, lying \\
        \bottomrule
    \end{tabular}
    \label{tab:keyword}
\end{table}

\subsection{Motion Coordinate Transformation}
To reduce redundancy in motion representation for diffusion model training, we transform each motion clip into a local coordinate system. 

Specifically, given the original motion $x_{0}$ in world coordinates, we will fetch the pose in the first frame $P_1=\{\theta_{1,global},\theta_{1,j=1:21},\tau_{1}\}$. Next, we can obtain the corresponding human joints $J=J_{1:K}$ from the human pose $P_1$, human shape $\beta$, and hand pose $\theta_h$ using the SMPL-X human model~\cite{pavlakos2019expressive}. In the SMPL-X human model, $J_{1:3}$ indicate the joints of the pelvis, left hip, and right hip. 

In the local coordinate system, the pelvis in the first frame should be positioned at the origin, thus we will first apply a translation $\mathbf{t}=-J_1$. Additionally, we need to apply a horizontal rotation $\mathbf{R}$ to ensure that the orientation of the human body in the first frame lie in the $yz$-plane ($y\ge 0$). When we only consider the local horizontal rotation for the initial human pose, we can set $\mathbf{z}_l = [0,0,1]$, $\mathbf{x}_l=\frac{\mathbf{n}}{||\mathbf{n}||}$ where $\mathbf{n}=[J_3[0]-J_2[0],J_3[1]-J_2[1],0]$ is the projection of $J_3-J_2$ on $xy$-plane, and $\mathbf{y}_l=\mathbf{z}_l\times\mathbf{x}_l$. At this point, the horizontal rotation can be formulated as
\begin{equation}
    \mathbf{R} = [\mathbf{x}_l^T,\mathbf{y}_l^T,\mathbf{z}_l^T]^{-1}.
\end{equation}
The full transformation takes the form 
\begin{equation}
    \mathbf{T}=
    \begin{bmatrix}
        \mathbf{R} & \mathbf{R}\mathbf{t}^{T} \\
        \mathbf{0} & 1
    \end{bmatrix}.
\end{equation}

We will apply the transformation $\mathbf{T}$ to the original motion $x_0$ in world coordinates, giving a new $x_0$ in local coordinates for diffusion model training. For motion synthesis, we will record the current transformation $\mathbf{T}$, and apply $\mathbf{T}^{-1}$ to the results to obtain the synthesized motion in world coordinates.

\subsection{Reward Design}
In our implementation, Eq. 5 in the main manuscript take the following reward functions as an implicit policy for scene-aware motion synthesis to pursue motion continuity, goal achievement, and plausible human-scene interaction.

\subsubsection{Historical Motion Consistency Reward.} To ensure continuity with historical motion $\tilde{P}_{1:\tilde{S}}$ when inferring future motion $\hat{P}_{1:S}$, the poses in the first $H$ frames of $\hat{P}_{1:S}$ should be consistent with the last $H$ frames in historical motion $\tilde{P}_{\tilde{S}-H+1:\tilde{S}}$. Here, we impose constraints on the body joints to maintain consistency between historical and synthesized future motions. Such reward can be formulated as:
\begin{equation}
    \mathcal{R}_{his} = \sum_{i=1}^{H} -|\hat{J}_i-\tilde{J}_{\tilde{S}-H+i}|. 
\end{equation}

\subsubsection{Smoothness Reward.} During motion synthesis, we should also ensure the motion smoothness, and such reward is necessary when interaction-based implicit policy is taken into the diffusion model. Thus, we introduce a reward for limited acceleration in case of abrupt pose changes:
\begin{equation}
    \begin{array}{rcl}
         \hspace{-0.1cm}\mathcal{R}_{acc} & \hspace{-0.3cm}= & \hspace{-0.3cm}\sum_{s=2}^{S-1}(||M_{m,s+1}+M_{m,s-1}-2M_{m,s}||_2\\
         & & \hspace{-0.3cm} \times \nu^2 - \epsilon_{acc}).
    \end{array}
\end{equation}
Here, $\epsilon_{acc}$ is the maximum tolerant acceleration.

\subsubsection{Goal Achievement Reward.} We also encourage the synthesized motion to achieve the goal, thus we take the goal position or pose joints in the form of $J_{goal}$ as the guidance. Here, we decide the frame $g$ that need to be controlled according to the distance and sampled speed (which is correlated with the action state). The goal achievement reward for the synthesized motion is defined as:
\begin{equation}
    \mathcal{R}_{goal} = -|\hat{J}_g - J_{goal}|.
\end{equation}

\subsubsection{Contact Reward.} The synthesized motion should keep in contact with the scene. For locomotion, at least one foot vertices should be in touch with the floor, thus the contact reward is define as:
\begin{equation}
    \begin{array}{rcl}
    \hspace{-0.1cm}\mathcal{R}_{cont} &\hspace{-0.3cm}=& \hspace{-0.3cm}\sum_{s=1}^{S}-ReLU(|\min_f(M_{f,s}[z]) - h_{floor}|\\
    & & \hspace{-0.3cm}-\epsilon_{cont}),
    \end{array}
\end{equation}
where $M_f$ represents the foot vertices, $\epsilon_{cont}$ is the tolerance for contact. As for other actions where other parts of the body should be in contact with the scene, we also utilize the scene SDF for guidance where reward can be defined as:
\begin{equation}
    \begin{array}{rcl}
    \hspace{-0.1cm}\mathcal{R}_{cont} &\hspace{-0.3cm}=& \hspace{-0.3cm}\sum_{s=1}^{S}-ReLU(|\min_m(f_{SDF}(M_{m,s}))|\\
    & &\hspace{-0.3cm}-\epsilon_{cont}).
    \end{array}
\end{equation}

\subsubsection{Non-Penetration Reward.} For the human-scene interaction in the synthesized motion, penetration should be avoided. We utilize the scene Signed Distance Function (SDF) as the guidance, and such reward function will encourage the generated human body move away from the interior space of the scene mesh. The non-penetration reward take the form of
\begin{equation}
    \mathcal{R}_{pene} = \sum_{s=1}^{S}\sum_{m}-ReLU(-f_{SDF}(M_{m,s})-\epsilon_{pene}).
\end{equation}
Here, $\epsilon_{pene}$ is the tolerance for slight penetration and we take SSM2 marker $M_m$ indicating $67$ body surface vertices to simplify the mesh like previous works~\cite{zhang2021we,zhao2023synthesizing}.

\subsubsection{Non-Skating Reward.} To avoid body skating during motion synthesis, we need to make sure that the velocity of contact body part to be close to $0$. Here, we decide to fetch all possible contact parts (\eg, feet during walking, gluteus and back during lying). Thus, the reward for non-skating can be formulated as follows:
\begin{equation}
    \begin{array}{rcl}
         \mathcal{R}_{skt} & \hspace{-0.3cm}=& \hspace{-0.3cm}\sum_{s=1}^{S-1}  -ReLU(\min_{c}(||M_{c,s+1}-M_{c,s}||_2) \\
         & & \hspace{-0.3cm} \times \nu - \epsilon_{vel}),
    \end{array}
\end{equation}
where $M_{c,s}$ indicates the contacted parts of mesh vertices at frame $s$, $\nu$ is the Frame Per Second (FPS), and $\epsilon_{vel}$ is the tolerance for minor skating.

Commonly, we set $\epsilon_{vel}=0.5$, $\epsilon_{pene}=0.03$, $\epsilon_{cont}=0.01$, and $\epsilon_{acc}=50$. For reward functions, $\lambda_{init}$ and $\lambda_{goal}$ are set to $1$, and $\lambda_{cont}$ is set to $10^{-1}$. For locomotion, $\lambda_{skt}$ is set to $10^{-3}$. For sitting, $\lambda_{skt}$, $\lambda_{pene}$, and $\lambda_{acc}$ are set to $3\times 10^{-4}$, $10^{-1}$, and $10^{-3}$. For lying, $\lambda_{pene}$ and $\lambda_{acc}$ are set to $3\times 10^{-2}$ and $10^{-3}$ respectively. All other coefficients are set to $0$.

\subsection{Motion Synthesis} The conditional motion diffusion model is designed to synthesize motion lasting $S=160$ frames within $T=10^{3}$ steps ($T_{inpaint}=50$). The first $K=22$ joints from SMPL-X model are selected to represent the body skeleton. Last $H$ (at most $H_{max}=10$) frames are taken as historical hints for motion blending. 

\subsection{Computing Infrastructure}
All experiments can be conducted on a RTX 3090 with 64G RAM. We also provide the explicit versions of software libraries in the supplementary code.

\section{More Visualization Results}
\begin{figure*}
    \centering
    \includegraphics[width=\linewidth]{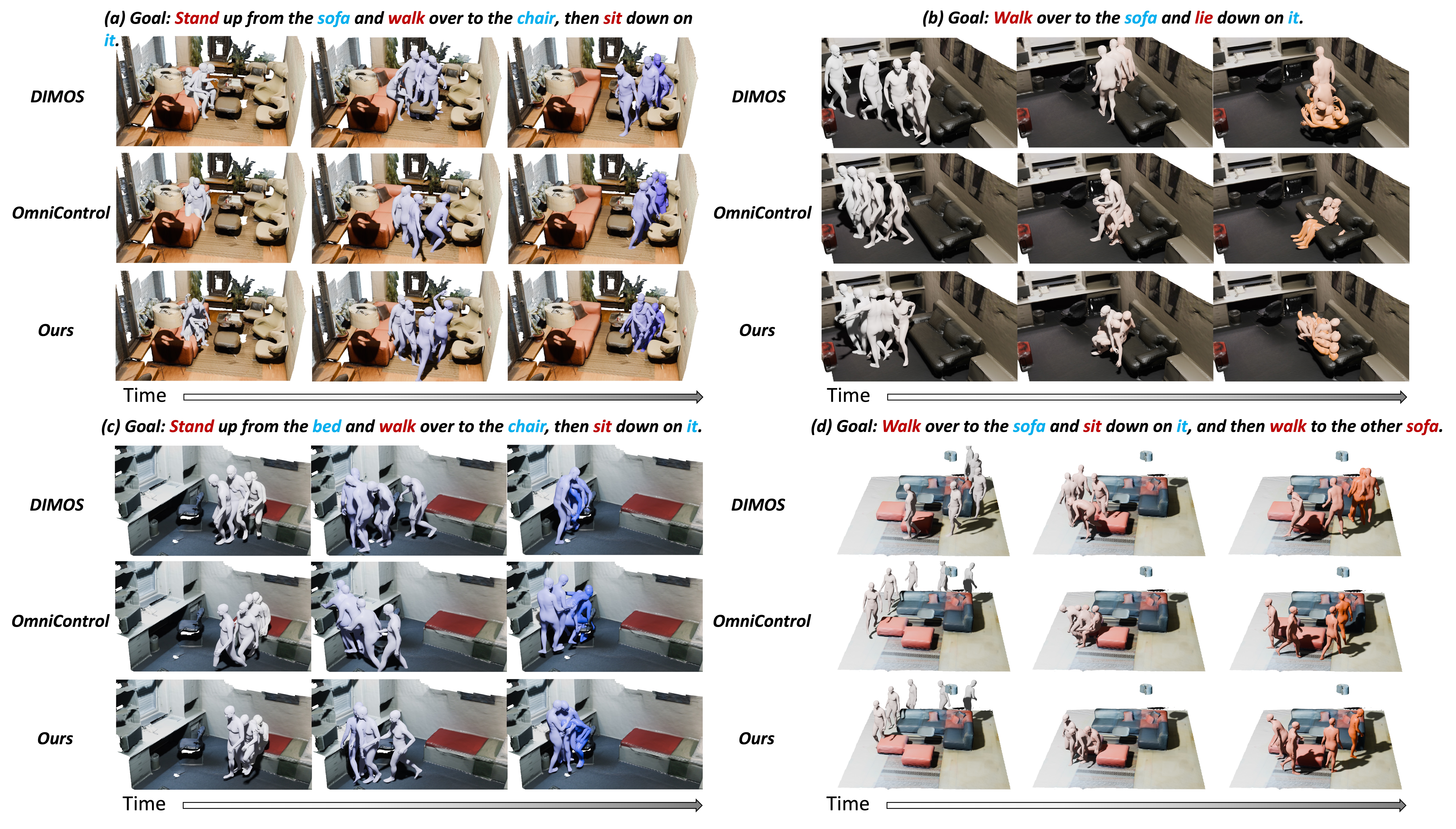}
    \caption{Additional visual comparison of different methods on motion synthesis in 3D scenes from PROX. }
    \label{fig:prox_supp}
\end{figure*}

\begin{figure*}
    \centering
    \includegraphics[width=\linewidth]{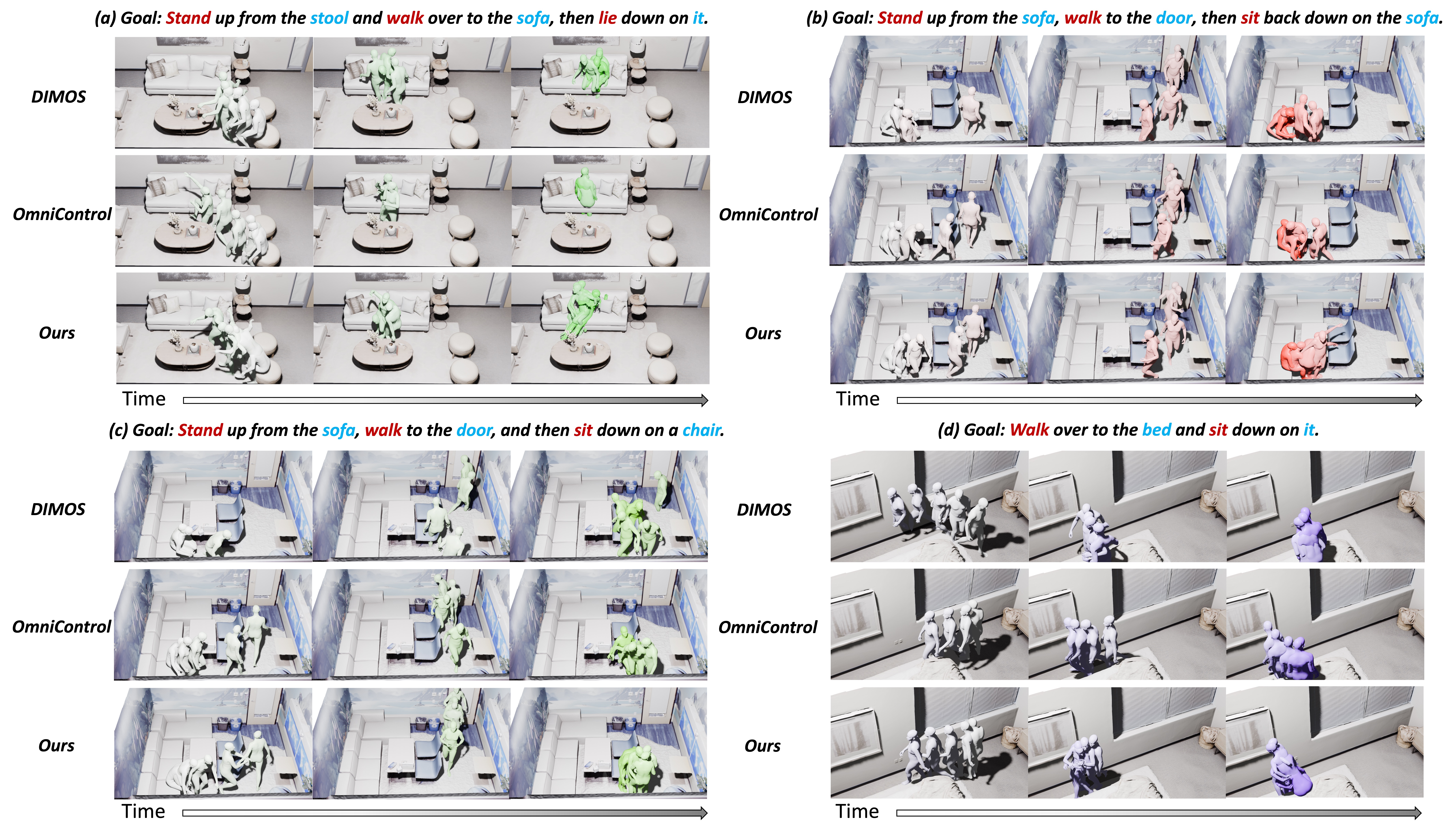}
    \caption{More visual results of synthesized motions given by compared methods in Replica scenes.}
    \label{fig:replica_supp}
\end{figure*}

In this section, we present more visualization results of
our method in scenes from PROX~\cite{hassan2019resolving} and Replica~\cite{straub2019replica}. In Fig.~\ref{fig:prox_supp}, we provide additional visual comparison of different methods in PROX dataset. In Fig.~\ref{fig:replica_supp}, we visualize more results of competitive methods on Replica dataset.

\subsection{Ablation Study}
\label{sec:ablation}

Here, we conduct more experiments to validate the proposed framework and prove our claims. 

\begin{figure}
    \centering
    \includegraphics[width=\linewidth]{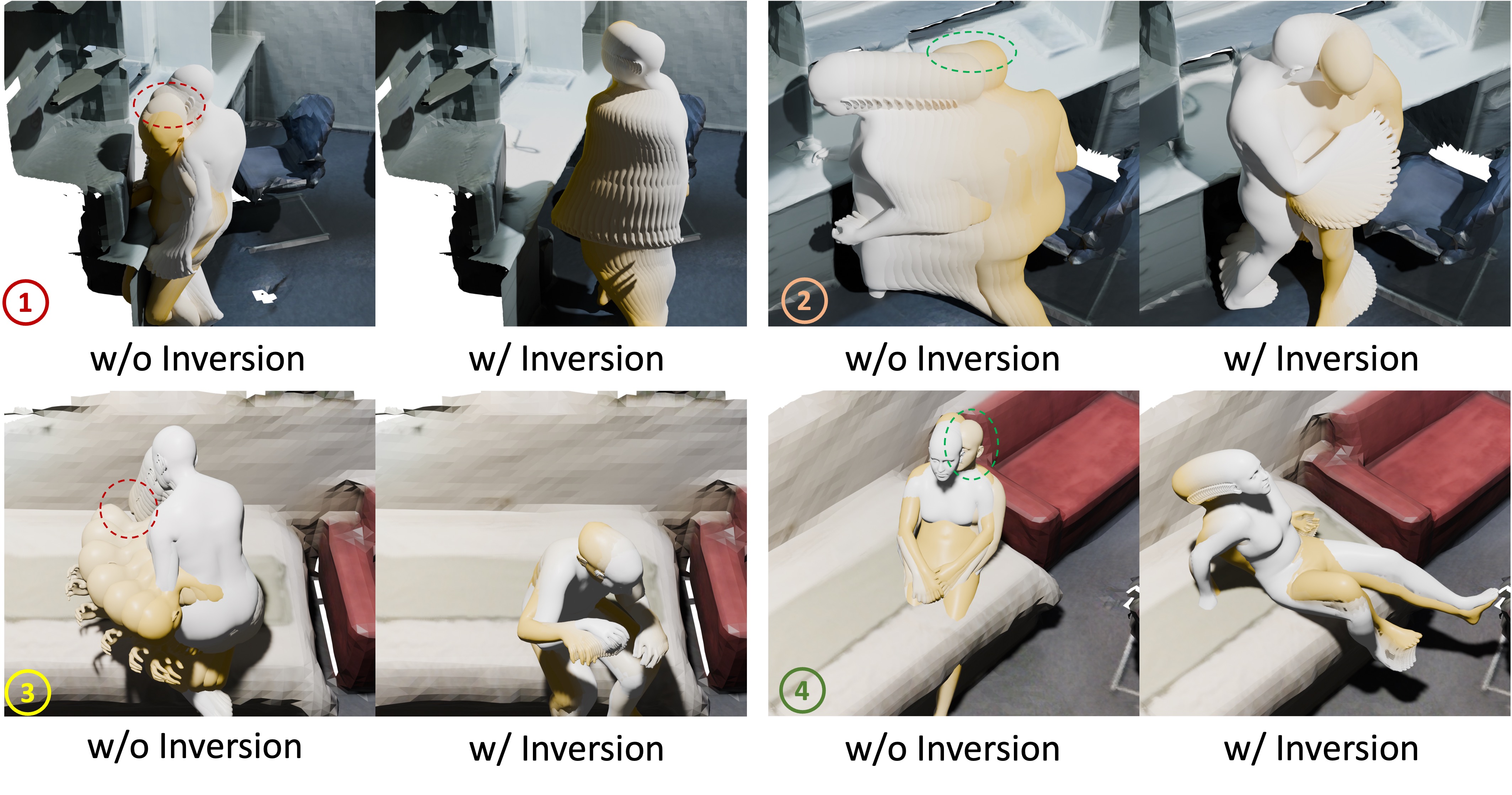}
    \caption{Comparison between two optimization strategy. For each pair, the left sub-figures show the results given by direct optimization, and the right sub-figures present the synthesized motions derived from optimization in GAN Inversion manner. Motions with obvious discontinuity are marked in red/green dashed circles.}
    \label{fig:inversion}
\end{figure}

\noindent\textbf{Direct Optimization v.s. GAN Inversion.}
In the implicit policy optimization, we propose to optimize $\mu_t$ via $\hat{x}_0^{\varphi}(\mu_t, t-1, c)$ rather than directly optimize $\mu_t$ as OmniControl~\cite{xie2024omnicontrol}. In this way, a better sample distribution centroid $\mu_t$ can be searched in GAN Inversion manner to satisfy the interaction in final synthesized motion $\hat{x}_0^\phi$. In addition, $\mu_t$ can be optimized as a whole instead of optimizing only one pose in one frame, thus can better keep motion continuity. Here, we compare the synthesized motions of these two strategies to prove the advantage of the proposed framework on final motion naturalness. Fig.~\ref{fig:inversion} illustrates the comparisons in a scene from PROX dataset, and we mark the synthesized motions with discontinuity in red/green circles. Without motion searching through GAN Inversion, the reward will directly guide the motion distribution centroid $\mu_t$, and we can see the sparse constrains in reward function will definitely make synthesized motions have abrupt changes.

\noindent\textbf{Inpainting.}
In order to keep consistent with the historical motion and better achieve the task goal, we decide to maintain the poses in key frames that need to be controlled in an inpainting manner. We synthesize motion in scenes with ShapeNet furniture with/without the keyframe pose inpainting, and judge the performance according to the goal achievement. We report the average distance to the desired goal position in Tab.~\ref{tab:inpainting}. As can be seen, the human can get closer to the destination when the motion are denoised with inpainting for human-scene interaction.

\begin{table}[htb]
    \centering
    \caption{Average distance to desired goal position for synthesized interaction motions with/without motion inpainting.}
    \begin{tabular}{lcc}
    \toprule
     & distance (sit $\downarrow$) & distance( lie $\downarrow$)  \\
    \midrule
     w/o inpainting & 0.14 & 0.11 \\ 
     w/ inpainting & \textbf{0.08} & \textbf{0.09}  \\ 
    \bottomrule
    \end{tabular}
    
    \label{tab:inpainting}
\end{table}

\section{Discussion} 
\label{sec:discussion}

According to the experiments in this paper, we could see entangling motion diffusion model and interaction-based implicit policy makes full utilization of the stochastic procedure in motion denoising, and can outperform current explicit policy method even without paired motion-scene data for training. This also indicates the proposed Diffusion Implicit Policy can generalize to diverse scenes as no specific scenes are required for training.

Even though, there are still some limitations in current method. As the implicit policy is introduced into motion denoising with limited gradient scale, there is still occasional collision between human and scene. Meanwhile, the motion style cannot be controlled currently, where action may be replaced by command in future work for easier style control.

\end{document}